\documentclass[conference]{IEEEtran}
\IEEEoverridecommandlockouts
\usepackage{cite}
\usepackage{amsmath,amssymb,amsfonts}
\usepackage{algorithmic}
\usepackage{graphicx}
\usepackage{textcomp}
\usepackage{xcolor}
\def\BibTeX{{\rm B\kern-.05em{\sc i\kern-.025em b}\kern-.08em
    T\kern-.1667em\lower.7ex\hbox{E}\kern-.125emX}}

\usepackage{booktabs}  
\usepackage{multirow} 
\usepackage{tabularx}

\newcommand{\modelname}{SHMoAReg}

\begin{document}

\title{SHMoAReg: Spark Deformable Image Registration via Spatial Heterogeneous Mixture of Experts and Attention Heads\\
}

\author{
    \begin{minipage}[t]{0.3\textwidth}
        \centering
        1\textsuperscript{st} Yuxi Zheng \\
        {\itshape 
        \small Institute of Science and Technology for Brain-inspired Intelligence \\
        \small Fudan University \\
        } 
        \small Shanghai, China \\
        \small \texttt{24110850037@m.fudan.edu.cn}
    \end{minipage}
    \hfill
    \begin{minipage}[t]{0.3\textwidth}
        \centering
        2\textsuperscript{nd} Jianhui Feng \\
        {\itshape 
        \small Institute of Science and Technology for Brain-inspired Intelligence \\
        \small Fudan University \\
        } 
        \small Shanghai, China \\
        \small \texttt{23210850006@m.fudan.edu.cn}
    \end{minipage}
    \hfill
    \begin{minipage}[t]{0.3\textwidth}
        \centering
        3\textsuperscript{rd} Tianran Li \\
        {\itshape 
        \small Institute of Science and Technology for Brain-inspired Intelligence \\
        \small Fudan University \\
        } 
        \small Shanghai, China \\
        \small \texttt{20110850023@fudan.edu.cn}
    \end{minipage}

    \\[3ex] 

    \hfill
    \begin{minipage}[t]{0.3\textwidth}
        \centering
        4\textsuperscript{th} Marius Staring \\
        {\itshape 
        \small Department of Radiology \\
        \small Leiden University Medical Center \\
        } 
        \small Leiden, The Netherlands \\
        \small \texttt{m.staring@lumc.nl}
    \end{minipage}
    \hfill
    \begin{minipage}[t]{0.3\textwidth}
        \centering
        5\textsuperscript{th} Yuchuan Qiao* \\
        {\itshape 
        \small Institute of Science and Technology for Brain-inspired Intelligence \\
        \small Fudan University \\
        } 
        \small Shanghai, China \\
        \small \texttt{yuchuanqiao@fudan.edu.cn}
    \end{minipage}
    \hfill 

    \thanks{*Corresponding Author.}
}

\maketitle


\begin{abstract}
Encoder-Decoder architectures are widely used in deep learning-based Deformable Image Registration (DIR), where the encoder extracts multi-scale features and the decoder predicts deformation fields by recovering spatial locations. 
However, current methods lack specialized extraction of features (that are useful for registration) and predict deformation jointly and homogeneously in all three directions.
In this paper, we propose a novel expert-guided DIR network with Mixture of Experts (MoE) mechanism applied in both encoder and decoder, named \textbf{\modelname}. 
Specifically, we incorporate Mixture of Attention heads (MoA) into encoder layers, while Spatial Heterogeneous Mixture of Experts (SHMoE) into the decoder layers.
The MoA enhances the specialization of feature extraction by dynamically selecting the optimal combination of attention heads for each image token. 
Meanwhile, the SHMoE predicts deformation fields heterogeneously in three directions for each voxel using experts with varying kernel sizes. 
Extensive experiments conducted on two publicly available datasets show consistent improvements over various methods, with a notable increase from 60.58\% to 65.58\% in Dice score for the abdominal CT dataset. 
To the best of our knowledge, we are the first to introduce MoE mechanism into DIR tasks.
\end{abstract}


\begin{figure*}[!t]
    \centering
    \includegraphics[width=0.8\linewidth]{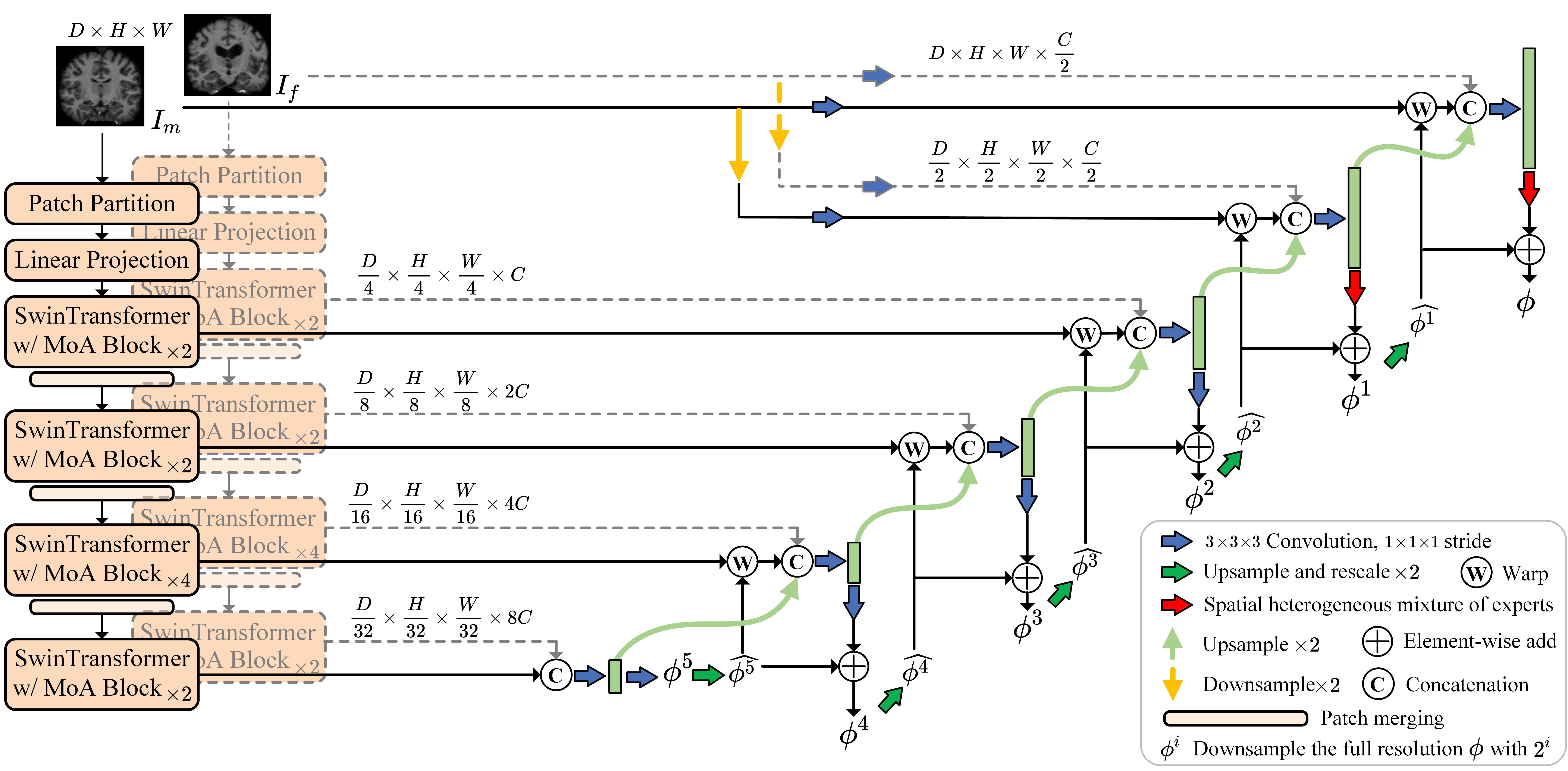}
    \caption{Overview of \textbf{\modelname}.
    The encoder has two parameter-sharing Swin Transformer backbones with Mixture of Attention heads (MoA) layers. The decoder employs the classic feature pyramid structure, where Spatial Heterogeneous Mixture of Experts (SHMoE) layers are introduced in generating the residual deformation fields at full and 1/2 resolutions.
    }
    \label{fig_overview}
\end{figure*}

\section{Introduction}\label{sec:intro}

Deep learning-based Deformable Image Registration (DIR) has many applications in computer-aided diagnosis and treatment. 
Until recently, most models in DIR~\cite{balakrishnan2019voxelmorph,che2023amnet,chen2022transmorph,tan2024groupmorph,chen2023transmatch,kang2022dual,wang2024recursive,shi2022xmorpher,hu2022recursive,meng2023non,qiao2021unsupervised,lv2022joint,chen2021vit,zheng2025gssd,mok2020large} primarily employ an encoder-decoder architecture, where the encoder extracts multi-scale features and the decoder recovers spatial location to predict deformation field. 
Convolutional Neural Network (CNN) layers are traditionally used~\cite{balakrishnan2019voxelmorph,hu2022recursive,kang2022dual,wang2024recursive,lv2022joint,mok2020large} to compose the encoder due to their capture of local features, but they struggle to model long-range dependencies. 
In contrast, Transformers~\cite{chen2022transmorph,chen2023transmatch,shi2022xmorpher,chen2021vit} become a popular choice as the encoder for better capturing global context. 
For the decoder, using a pyramid structure~\cite{hu2022recursive,lv2022joint,mok2020large,meng2023non,wang2024recursive,tan2024groupmorph}, which progressively generates deformation fields from coarse to fine, has become a widely popular approach in recent years.
Despite the substantial success of existing encoder-decoder based works, they still exhibit some limitations:

\textbf{\textit{Firstly}}, the feature extraction in the encoder lacks explicit specialization. 
Customizing an optimal combination of attention heads for each image token in Transformers to match its specific role for deformation learning is rarely explored.
However, different heads emphasize different pair-wise attentions and visual dependencies~\cite{li2023does}, especially when capturing multi-scale features across different resolution layers and modeling spatial correlations within the same resolution layer. The different contributions of each attention head should be considered.
\textbf{\textit{Secondly}}, current pyramid-based decoders~\cite{tan2024groupmorph,wang2024recursive} generally perform the fusion or generation of multiple subfields using the same convolutional kernel size for all three directions. Such direction-homogeneous predictions neglect the inherent structural properties in specific orientations of medical images, especially when dealing with anisotropic voxel spacings. Therefore, it is crucial to learn deformations at a fine-grained per-direction level to better model the localized and directional anatomical variations.

Some researchers had introduced Mixture of Experts (MoE) into deep learning models to improve feature specialization~\cite{liu2024learning}, particularly within encoder-decoder architectures for medical image segmentation~\cite{jiang2024m4oe,ou2022patcher,novosad2024task,wang2024sam,chen2024low}, where different experts are used to learn modality-specific segmentation~\cite{jiang2024m4oe,novosad2024task} or downstream organ-specific segmentation~\cite{wang2024sam,chen2024low}.
However, no previous work has applied the MoE mechanism to the encoder-decoder structure for learning spatial correspondences in deformable image registration.
Integrating MoE mechanism into DIR networks confronts the following \textbf{\textit{major challenges}}:
\begin{itemize}
    \item \textbf{\textit{How to}} extract more specialized features in different resolutions of encoder layers?
    \item \textbf{\textit{How to}} perform spatial per-voxel, per-direction level differentiated deformation learning in the decoder?
\end{itemize}

 In this paper, we address the above challenges and design a novel expert-guided registration network with MoE mechanism applied in both encoder and decoder, named \textbf{\modelname}. We highlight the main contributions as follows:
\begin{itemize}
    \item \textbf{We incorporate the Mixture of Attention heads (MoA) into the encoder layers for specialized feature extraction.} 
     MoA enables every image token to dynamically select the optimal attention heads combination from a larger subset when extracting features. 
    \item \textbf{We introduce Spatial Heterogeneous Mixture of Experts (SHMoE) for differentiated deformation learning.}
     We assign experts with different kernel sizes to specialize the magnitude of deformation along each direction for each voxel. 
     \item \textbf{We validated the effectiveness and generalization of our approach on public datasets.}
     Experiments conducted on brain MR and abdominal CT datasets show stable improvements compared to both CNN- and Transformer-based encoder-decoder models across three mainstream architectures (U-shape, Cascade, Pyramid).
\end{itemize}

\section{Methods}\label{sec:methods}
Given a pair of 3D moving and fixed images $\{I_{m}, I_{f}\}$, our objective is to estimate a deformation field $\phi$ such that the warped image $I_{w} = I_{m} \circ \phi$ can be aligned with $I_{f}$. 
Fig.~\ref{fig_overview} illustrates the overview of the encoder-decoder framework of our proposed \textbf{\modelname}. 
Specifically, the encoder comprises two shared-parameters Swin Transformer backbones as in~\cite{chen2022transmorph}, which respectively extract features from $I_{m}$ and $I_{f}$. 
To tailor a combination of attention heads suitable for each token's functionality, we replace the Window/Shifted Window-based Multi-head Self-Attention (W/SW-MSA) layers in each Swin Transformer Block with Mixture of Attention heads (MoA) layer~\cite{zhang2022mixture}. 
The MoA dynamically selects the attention heads combination from a larger subset for each image token to extract features specialized for registration.

The decoder follows the classic and effective feature pyramid structure~\cite{meng2023non,wang2024recursive}, where features from each level are warped using the deformation field predicted by the former layer, learning the final deformation field $\phi$ in a coarse-to-fine manner. 
Notably, when generating the residual deformation fields at full and 1/2 resolution, we introduce Spatial Heterogeneous Mixture of Experts (SHMoE) layers to differentiate the deformation prediction for each direction (x/y/z).
The SHMoE layer comprises a set of experts and a routing gate.
The experts are designed as convolution layers with different receptive fields (e.g., 1x1x1, 3x3x3, and 5x5x5 kernel sizes), which allows for a heterogeneous selection of deformations across varying spatial scales. 
Meanwhile, the routing gate selects the top-k expert (k=1) at a fine-grained per-voxel and per-direction level, which is particularly effective for handling anisotropic voxels in dense DIR tasks.

Our \modelname~is supervised by a similarity loss and a regularization loss in common registration tasks.
Furthermore, we introduce a binary cross-entropy loss as our Routing Classification (RC) loss between the routing gate tensor $T$ and constructed labels $Y$ to determine if the expert selection for each voxel in SHMoE is 'correct'.
The total loss of our \textbf{\modelname} can be formulated as:
\begin{align}
    \mathcal{L}_{total}= \mathcal{L}_{sim}(I_{w},I_{f} ) + \lambda_{r} \mathcal{L}_{reg}(\phi ) + \lambda_{rc} \mathcal{L}_{rc}(T, Y), 
    \label{L_TOTAL}
\end{align}
where $\lambda_{r}$ is the weight of the regularization term and $\lambda_{rc}$ is the weight of the routing classification loss.

\begin{table*}[t]
  \centering
  
  \caption{
  Quantitative results over comparison of encoder-decoder-based methods and our method on two datasets. Symbol $\!*\!$ marks results where {\color{red}\modelname} significantly outperforms the {\color{blue}second-best} method ($p\!<\!0.05$, Wilcoxon signed-rank test). 
  }
    {\fontsize{8pt}{9pt}\selectfont  
    
    \begin{tabular}{cccccccccc}
    
    \toprule

     \multirow{2}{*}{Type} &\multirow{2}{*}{Methods} & \multicolumn{3}{c}{OASIS} & \multicolumn{3}{c}{BTCV (w/o pre-affine)} & Time & Params \\
    \cmidrule(lr){3-5} \cmidrule(lr){6-8} & & DSC$_{(\%)}$ & ASSD$_{(\mathrm{mm})}$ & $\left | \! J_{\phi} \! \le \!0 \!\right |_{(\%)}$ & DSC$_{(\%)}$ & ASSD$_{(\mathrm{mm})}$ & $\left | \! J_{\phi} \! \le \!0 \!\right |_{(\%)}$ &(s) &(M) \\

    \midrule

    Initial &/ & 23.30${\pm2.95}$ & 3.11${\pm0.53}$ & / & 26.31${\pm10.70}$ & 7.62${\pm2.21}$ & / & / & / \\

    \midrule

    \multirow{2}{*}{U-shape} &VM & 76.54${\pm3.63}$ & 0.42${\pm0.14}$ & 0.07 & 56.71${\pm9.22}$ & 3.25${\pm0.99}$ & 0.02 & 0.02 & 0.33 \\

    &TM & 77.88${\pm2.90} $ & 0.38${\pm0.11} $ & 0.06 & 55.75${\pm9.48} $ & 3.30${\pm1.02} $ & 0.08 & 0.04 & 46.69 \\
    
    \midrule

    \multirow{2}{*}{Cascade} &2casVM & 78.40${\pm2.40}$ & 0.37${\pm0.09}$ & 0.06 & {\color{blue}60.58${\pm7.93}$} & 2.85${\pm0.76}$ & 0.08 & 0.06 & 0.65 \\

    &2casTM & {\color{blue}79.35${\pm2.10} $} & {\color{blue}0.33${\pm0.08} $} & 0.07 & 59.57${\pm9.55} $ & 2.97${\pm0.87} $ & 0.13 & 0.10 & 93.38 \\

    \midrule

    \multirow{2}{*}{Pyramid} &N-T & 78.99${\pm1.87}$ & 0.35${\pm0.07}$ & 0.06 & 60.25${\pm7.64}$ & {\color{blue}2.81${\pm0.69}$} & 0.06 & 0.08 & 5.62 \\

    &RDP & 79.18${\pm1.94} $ & 0.35${\pm0.07} $ & $\!<\!0.01\!$ & 60.14${\pm8.17} $ & 2.89${\pm0.80} $ & $\!<\!0.01\!$ & 0.19 & 8.92 \\

    \midrule

    \multirow{2}{*}{MoE} &\modelname & {\color{red}79.95$^{*}\!{\pm1.88}\!$} & {\color{red}0.31$\!{\pm0.07}$} & 0.43 & {\color{red}65.58$^{*}\!{\pm7.29}$} & {\color{red}2.46$^{*}\!{\pm0.58}$} & 0.50 & 0.09 & 18.88 \\

    &\modelname$_{\textit{diff}}$ & {\color{red}79.86 ${\pm2.02}\!$} & {\color{red}0.32$\!{\pm0.08}$} & $\!<\!0.01\!$& {\color{red}64.15$^{*}\!{\pm7.46}$} & {\color{red}2.59$^{*}\!{\pm0.70}$} & $\!<\!0.01\!$ & 0.09 & 18.88 \\

    \bottomrule

    \end{tabular}
    }
  \label{table1: main result}
\end{table*}
\begin{figure*}[!tbp]
    \centerline{\includegraphics[width=0.8\textwidth]{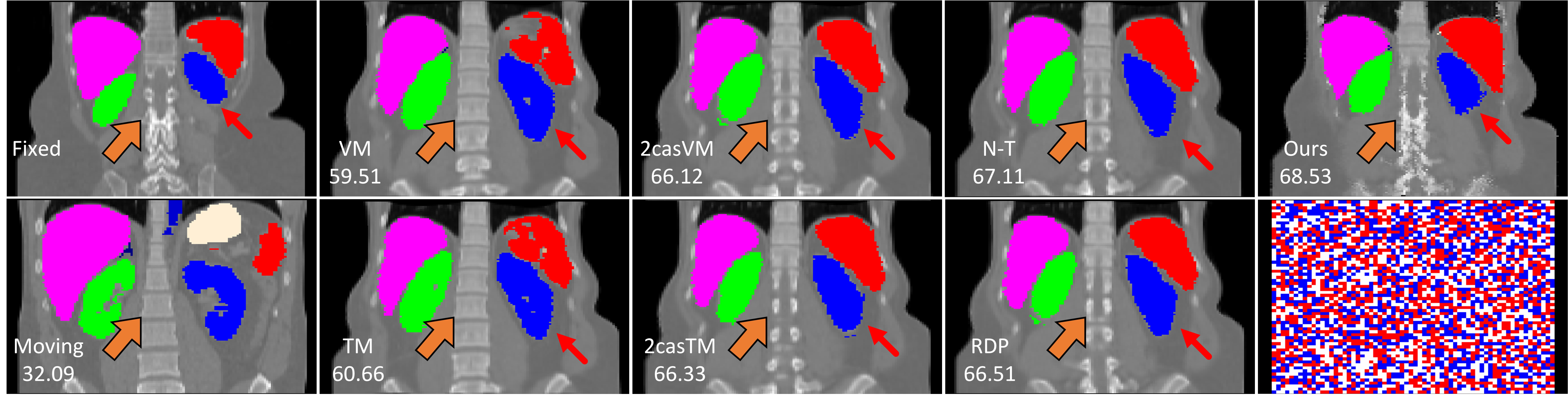}}
    \caption{Visualization of the warped images produced by comparison methods and our \modelname~ (Dice scores shown in the bottom left). For this slice in the x-direction, the expert ID selected at each voxel is also visualized (red, white, and blue represent 3 different experts).
    }
	\label{fig_visual_result}
\end{figure*}

\section{Experiments}\label{sec:experiment}
\subsubsection{Implementation}
We evaluate all methods on the brain MR (OASIS~\cite{marcus2007open}) and abdominal CT (BTCV~\cite{landman2015miccai} without pre-affine) datasets with standard pre-processing steps, including center cropping, resampling to $128\! \times \! 128\! \times \!128$ and intensity norm to [0,1]. 
For OASIS (isotropic voxel spacing with 1mm), we randomly select 350 (350$\!\times\!$349 pairs), 10 (10$\!\times\!$9 pairs), and 11 (11$\!\times\!$10 pairs) volumes for training, validation, and test sets, respectively. 
For BTCV ($3\! \times \! 3\! \times \!2$ mm anisotropic voxel spacing), the divisions are 35 (35$\!\times\!$34 pairs), 5 (5$\!\times\!$4 pairs), and 10 (10$\!\times\!$9 pairs), respectively.
35 region labels in the brain and 4 primary organ labels (the liver, the spleen, the right kidney and the left kidney) are used for evaluation, using: the Dice score of the segmentation maps (DSC$_{(\%)}$)~\cite{balakrishnan2019voxelmorph}, the Average Symmetric Surface Distance (ASSD)~\cite{taha2015metrics} between segmentation maps, the percentage of voxels with non-positive Jacobian determinant ($\left | \! J_{\phi} \! \le \!0 \!\right |_{(\%)}$)\cite{mok2020large}, GPU inference time and model parameters.
All methods are implemented in PyTorch~\cite{paszke2017automatic}, using an NVIDIA L40 GPU with 48GB. 
We choose MSE loss as our similarity loss for all datasets and employ Adam~\cite{kingma2014adam} optimizer with a fixed learning rate of 1e-04 and a batch size of 1 to train our model for 100,000 iterations. 
The parameter $\lambda_{rc}$ is set as 0.001 for all datasets, while $\lambda_{r}$ is set as 0.01 for OASIS and 0.1 for BTCV based on the validation results.
The encoder's SwinTransformer backbone is set the same as TransMorph-small~\cite{chen2022transmorph}.

\subsubsection{Comparison}
We compare \modelname~with three types of encoder-decoder methods: (1) U-shape: VoxelMorph (VM)~\cite{balakrishnan2019voxelmorph}, TransMorph (TM)~\cite{chen2022transmorph}; (2) Cascaded: 2casVM and 2cas TM; (3) Pyramid: NICE-Trans (N-T)~\cite{meng2023non} and RDP~\cite{wang2024recursive}.
All methods were implemented using their official releases with default parameters.

\section{Results}\label{sec:Result}
As shown in Table~\ref{table1: main result}, our method achieves a Dice score of 65.58\% and an ASSD of 2.46mm on the challenging BTCV dataset with anisotropic voxel spacings, outperforming the second-best method by margins of 5.0\% and 0.35mm, respectively. These results demonstrate the effectiveness of incorporating Spatial-Heterogeneous deformation learning into registration models.
Similar trends are also observed on the isotropic OASIS dataset.

To explore the advantages of \modelname~in cases where no folding in the deformation field (i.e. $\left | \! J_{\phi} \! \le \!0 \!\right |_{(\%)}$ approximately equals to 0), we also present the results of diffeomorphic variant \textbf{\modelname$_{\textit{diff}}$} in Table~\ref{table1: main result}. 
Under the condition of preserving the topology, \modelname$_{\textit{diff}}$ achieves a 3.57\% higher Dice score and a 0.22mm lower ASSD on the BTCV dataset compared to the second-best method. 
Similar improvements are also observed in the OASIS datasets. These results demonstrate the effectiveness of the MoE mechanism, confirming that the improvements in registration accuracy are not accompanied by a compromise in the smoothness of the deformation field.

Fig.~\ref{fig_visual_result} illustrates the warped images from different methods on one example subject from the BTCV dataset. Notably, our method achieves the closest alignment with the fixed image on the left kidney (blue label) and uniquely aligns it with the vertebral body, which was challenging to register without pre-affine.
We also visualize the expert ID selected for each voxel in the x-direction for the corresponding slice, showing SHMoE's spatial heterogeneous expert selection at a per-voxel level. 
It appears sufficiently fine-grained to resemble noise because each SHMoE is designed to learn the residual deformation. Even voxels in adjacent regions may exhibit different patterns of residual deformation due to the accumulative effect of prior deformation learning, thereby causing the selection of different kernel-specific experts.

Conclusively, the substantial quantitative enhancements observed in both isotropic and anisotropic datasets, the robust diffeomorphic performance, and the fine-grained qualitative alignments all serve to provide comprehensive validation for our proposed MoE-based \modelname.

\section{Conclusions and Discussion}\label{sec:conclusions}
In this paper, we propose a novel expert-guided registration network with MoE mechanism applied in both encoder and decoder, named \textbf{SHMoAReg}.  
The encoder's MoA enables specialized feature extraction in different resolution layers, while the decoder's SHMoE empowers spatial per-voxel, per-direction level heterogeneous deformation learning.
In conclusion, we have successfully introduced the MoE mechanism into DIR tasks, and our \modelname~demonstrates promising results on various datasets.
More heterogeneous fine-grained experts are left to be explored in the future, especially for multimodal and low-quality images.

Due to space limitations, detailed ablation studies on the impact of MoE mechanism and the analysis of model interpretability and specialization will be presented in future work.

\section*{Acknowledgments}
This work is supported by the National Natural Science Foundation of China under Grant 82102002.

\bibliographystyle{IEEEtran}
\bibliography{reference}

\end{document}